# MTLComb: multi-task learning combining regression and classification tasks for joint feature selection

Han Cao[1], Sivanesan Rajan[2,3], Bianka Hahn[4], Ersoy Kocak[2,3], Daniel Durstewitz[1], Emanuel Schwarz[2,3]*, Verena Schneider-Lindner[4]*

1 Department of Theoretical Neuroscience, Central Institute of Mental Health, Medical Faculty, Heidelberg University, Germany

2 Department of Psychiatry and Psychotherapy, Central Institute of Mental Health, Medical Faculty, Heidelberg University, Mannheim, Germany

3 Hector Institute for Artificial Intelligence in Psychiatry, Central Institute of Mental Health, Medical Faculty Mannheim, Heidelberg University, Mannheim, Germany

4 Department of Anesthesiology and Surgical Intensive Care Medicine, Medical Faculty Mannheim, Heidelberg University, Theodor-Kutzer-Ufer 1-3, 68167, Mannheim, Germany

## Abstract

Multi-task learning (MTL) is a learning paradigm that enables the simultaneous training of multiple communicating algorithms. Although MTL has been successfully applied to ether regression or classification tasks alone, incorporating mixed types of tasks into a unified MTL framework remains challenging, primarily due to variations in the magnitudes of losses associated with different tasks. This challenge, particularly evident in MTL applications with joint feature selection, often results in biased selections. To overcome this obstacle, we propose a provable loss weighting scheme that analytically determines the optimal weights for balancing regression and classification tasks. This scheme significantly mitigates the otherwise biased feature selection. Building upon this scheme, we introduce MTLComb, an MTL algorithm and software package encompassing optimization procedures, training protocols, and hyperparameter estimation procedures. MTLComb is designed for learning shared predictors among tasks of mixed types. To showcase the efficacy of MTLComb, we conduct tests on both simulated data and biomedical studies pertaining to sepsis and schizophrenia.

## Introduction

Multi-task learning (MTL) is a machine learning paradigm that allows simultaneous learning on different yet related tasks. By leveraging the inherent relatedness among tasks, MTL facilitates the acquisition of a shared representation, potentially enhancing the model's generalizability. Since its inception[1], MTL has been widely applied in biomedicine[2], vision[3], audio[4] and natural language processing[5], as well as internet engineering[6]. In the early stages of development, one of the primary motivations for employing MTL was the challenge of data scarcity. For instance, in molecular biology, the sample size is often considerably smaller than the number of features due to the high cost of data acquisition. This makes coefficient estimation difficult, given the high model variances. MTL, in this scenario, mitigates model estimation variance by sharing prediction-relevant information among tasks. In the current 'big data' era, MTL is recognized for its ability to efficiently process larger volumes of data from multiple modalities[7], showing an advanced learning efficiency[7] and prediction accuracy[3].

MTL involving mixed types of learning tasks presents challenges that have been explored across various fields. In computer vision, for instance, the scene understanding algorithm[3] engages in joint learning of multiple regression (e.g., semantics prediction) and classification tasks (e.g., geometry prediction). Likewise, in dynamic system reconstruction applications, such systems can be inferred from the time series of multiple data modalities[8,9] (e.g., continuous data and binary data). This process includes the joint fitting of regression and categorical models using multimodal time series. Similarly, in biomedicine, predictive models of Alzheimer's disease have considered the joint prediction of diagnostic labels and the cognition scores[10,11]. These approaches have demonstrated significant predictive capacity and mechanistic interpretability. To achieve this, an MTL model allowing joint regression and classification becomes essential. To circumvent the necessity for such joint learning, the continuous outcomes are, in practice, often binarized given a certain threshold, and classification-based MTL is subsequently employed. As highlighted in [3], the challenge in designing MTL with mixed types of learning tasks lies in balancing the different magnitudes or scales resulting from various types of losses. For example, the study by [list author et al.][3] showed that different types of losses lead to varying magnitudes of task noise, confusing the model parameter optimization. The same study recommended a loss weighting scheme to balance losses of different types.

The loss weighting scheme is a model training method employed to balance individual loss functions for different tasks[7]. In the early stages of MTL, uniformly or constantly applied weights did not consider the specificity of different types of losses[12,13]. In the recent years, there has been a growing focus on the study of learnable weights due to their adaptability to the data. This is achieved by incorporating the weights as variables into the machine learning architecture and estimating them through optimization[7]. Such consideration for weights may include estimation uncertainty[3], learning speed[14] and prediction accuracy[15]. However, these approaches might complicate the training procedure because they involve the iterative estimation of weights and models. Therefore, an analytical solution that can determine the weights piror to model training would have significant utility. To address this, we developed MTLComb, which can analytically determine the optimal weights for regression and classification losses in linear MTL.

MTLComb is based on linear Multi-Task Learning (MTL) for the joint learning of regression and classification tasks. Here, our primary focus was on the selection of features relevant across both task types[16]. Performing such joint feature selection entails identifying a subgroup of features that is identical and optimally predictive in both regression and classification tasks. It is important to note that the theory underlying MTLComb is broadly applicable to other linear MTL approaches, such as low-rank[17] and network-constranted[18] approaches. As depicted in **Figure 1**, the challenge of joint feature selection arises from the lack of alignment of the feature selection principles, also known as regularization paths[19], between classification and regression tasks. This misalignment depends on the divergent magnitudes of losses, leading to biased feature selection. For instance, in **Figure 1**, when $\lambda = 0.8$, seven features are selected for regression tasks, while none are selected for classification tasks. Our approach, MTLComb, mitigates this issue by employing a simple and provable losses weighting scheme. We demonstrate the effectiveness of MTLComb through two biomedical case studies focusing on sepsis and schizophrenia.

Sepsis is the leading cause of mortality in Intensive Care Units (ICU)[20] and critical illness worldwide[21]. It is caused by the dysregulated host response to infection and organ dysfunction, and has been associated with high risk of complications, e.g., ICU death, longer hospital stays and higher medical costs[22,23]. Early detection of sepsis plays a crucial role in improving patient survival rates[24] and reducing medical expenses. As a response, a research line has emerged to predict various clinical outcomes of sepsis, such as diagnosis[25], ICU motality[23] and ICU stay[26], leveraging early clinical features — typically referring to patient data upon ICU admission. However, individual outcome (e.g., diagnosis of sepsis) prediction model may not fully capture the underlying risk patterns associated with sepsis. In this case study, we employ MTLComb to discern sepsis risk factors that can simultaneously predict diagnosis (classification task), kidney function (regression tasks), and metabolic measurements (regression task). This choice is grounded in the understanding that the changes in metabolic variation and kidney functioning accompanying sepsis onset are driven by the same biological processes. We hypothesize that this multi-task learning formulation will yield a more predictive and interpretable model.

Schizophrenia, a severe mental illness, impacts approximately 24 million people worldwide, contributing to a massive societal and economic burden[27]. Neuroimaging studies have shown that schizophrenia is associated with an accelerated aging of the brain. These studies are based on the concept of brain age — are inferred from brain-structural or other data in contrast to chronological age — to quantify a potential acceleration of aging. This approach has uncovered evidence of age-dependent changes in brain structure[28], epigenetic data[29], and gene expression[30]. To further explore age dependent effects in schizophrenia, we explore the joint prediction of age (regression task) and diagnosis (classification task) as an MTLComb case study. We assembled two cohorts comprising brain expressions from individuals diagnosed with schizophrenia and healthy controls, designated as the discovery and validation cohorts. On the discovery cohort, MTLComb model was employed to identify age-dependent genes that simultaneously serve as predictive markers for schizophrenia diagnosis. Subsequently, we assessed the reproducibility of these markers using a validation cohort and explored the associated pathways for biological interpretation.

The following sections present the MTLComb method, a simulation data analysis demonstrating its prediction performance and feature selection accuracy compared to other approaches, and two case studies of real data analysis.

## Methods

### Intuition of MTLComb

**Figure 1** demonstrates the major challenge of linear MTL with mixed types of tasks for joint feature selection, as well as the conceptual basis of MTLComb. The feature selection principle is quantified as the regularization path[19] that is parameterized by a hyperparameter λ. Each value of λ determines a subset of selected features and their coefficients. A higher value of λ is associated with a smaller number of selected features. Due to the different magnitudes of losses, the regularization paths of regression and classification tasks are not aligned, as shown in **Figure 1**, leading to a biased joint feature selection. For example, the regression tasks dominate entirely the joint feature selection when λ ∈ $[0.6,1.2]$. In this manner, classification tasks can not have any selected features. The primary aim of MTLComb is to align the regularization paths of different tasks to the same root.

The intuition underlying MTLComb is simple. Consider a least-square loss of a regression problem weighted by $\alpha$, $\min_{w} \alpha||Y - Xw||_2^2$, where the solution is $w = \alpha(X^T X)^{-1} XY$. This means, the maganitude of $w$ can be adjusted by $\alpha$, leading to a movable regularization path. Extending this intuition to multiple types of losses, we are allowed to find optimal weights of different losses, aligning the feature selection principles. We proved in **Proposition 1** that the constant weights used in MTLComb are optimal. Further details of the algorithm are provided in the following section.

### Modeling, optimization and algorithms

The formulation of the MTLComb is shown in (1)

$$\min_{W} 2 \times \mathrm{Z}(W) + 0.5 \times \mathrm{R(W)} + \lambda||W||_{2,1} + \alpha||WG||_2^2 + \beta||W||_2^2 \quad (1)$$

where $\mathrm{Z}(W) = \sum_{i=1}^{c} \frac{1}{N_i} \log(1 + e^{-Y^{(i)}(X^{(i)}w^{(i)})})$ , $\mathrm{R}(W) = \sum_{i=c+1}^{t} \frac{1}{N_i} ||Y^{(i)} - X^{(i)}w^{(i)}||^2$ , $W = [w^{(1)}, \dots w^{(c)}, \dots w^{(t)}]$ and $G = \mathrm{diag}(t) - \frac{1}{t}$.

$\mathrm{Z}(W)$ is the logit loss to fit the classification tasks, and $\mathrm{R}(W)$ is the least-square loss to fit the regression tasks. $X = \left\{X^{(i)} \in \mathbb{R}^{N_i \times p}: i \in \{1, \dots, c, \dots t\}\right\}$, refers to the feature matrices of $t$ tasks, where $p$ features are consistent across tasks. $Y = \left\{Y^{(i)} \in \mathbb{R}^{N_i \times 1}: i \in \{1, \dots, c, \dots t\}\right\}$ describes the outcome lists associated with $c$ classification and $t - c$ regression tasks. $W \in \mathbb{R}^{p \times t}$ is the coefficient matrix that needs to be estimated, where each column represents the coefficient vector of each task, and each row $w^{(j)} \in \mathbb{R}^{t \times 1}$ consists of the coefficients of feature $j$. $||W||_{2,1} = \sum_{j=1}^{p} \sqrt{||w^{(j)}||_2^2}$ is a sparse penalty term to promote the joint feature selection[16]. $||WG||_2^2$ is the mean-regularized term[31,32] to promote the selection of features with similar cross-task coefficients. $||W||_2^2$ aims to select the correlated features and stabilize the numerical sulutions[33]. $\{\lambda, \alpha, \beta\}$ is the hypermeter set, which controls the strength of the penalty. Here, $\lambda$ is selected by cross-validation, and users select the rest as constant prior. We weight $\mathrm{Z}(W)$ by 2 and $\mathrm{R}(W)$ by 0.5. This simple weighting scheme made the regularization paths consistent.

To solve the objective shown in (1), we adopt the accelerated proximal gradient descent method to approximate the solution[34], which features a "state-of-art" algorithmic complexity of $O(1/k^2)$. The derivation of the optimization procedures, the relevant algorithms are summarized in the supplementary methods.

**Regularization path estimation**

The regularization path, exemplified by the Lasso[19], illustrates the feature selection principle. MTLComb's central function is to estimate the complete regularization path, representing a series of models indexed by a sequence of $\lambda$ (a spectrum of sparsity levels). Accurately determining the $\lambda$ sequence is crucial to capture the highest likelihood while avoiding unnecessary explorations. Inspired by glmnet[33], we estimated the $\lambda$ sequence from the data in three steps. First, we estimate the largest $\lambda$ (referred to as 'lam_max') in the sequence, leading to nearly zero coefficients. Second, we calculate the smallest $\lambda$ in the sequence using lam_max (e.g., 0.01 * lam_max). Third, we interpolate the entire sequence on the log scale. Calculating lam_max with mixed losses poses a challenge, as the optimal estimate for the least squares loss being incompatible with that for the logit loss, as shown in **Figure 1**. Using lam_max based on either loss alone could introduce bias in biomarker identification. **Proposition 1**, detailed in the supplementary methods, demonstrates that a consistent lam_max for both classification and regression tasks can be determined by weighting the regression and classification losses, as shown in formulation (1). The proof of Proposition 1 is available in the supplementary methods, and the algorithm for estimating the entire regularization path is summarized therein.

**Proposition 1**. By weighting $\mathrm{Z}(W)$ with 2 and $\mathrm{R}(\mathrm{W})$ with 0.5, we can quantify an identical form of lam_max satisfying both regression and classification tasks

## Simulation analysis

In this analysis, we compare various approaches in the context of mixed regression and classification tasks. Multiple classification and regression tasks are generated based on the same set of randomly generated ground truth features. The objective is to assess the prediction performance of these approaches and the recovery rate (joint feature selection accuracy) of the ground truth features in the context of mixed task types. The detailed construction protocol for simulation data is provided in the supplementary methods. Prediction performance is quantified as the (pseudo-) explained variance for classification and regression tasks. Additionally, we varied the ratio of $\frac{\text{subject number}}{\text{feature number}}$ from 0.1 to 0.8 to simulate high- and low-dimensional data problems. To account for sampling variation, we conducted the analysis five times and averaged the results.

The comparison includes two types of approaches: 1) MTL approaches and 2) meta-analysis of individual machine-learning models. In MTL approaches, alongside MTLComb, we consider a binarization approach (MTLBin) as a baseline. MTLBin involves binarizing continuous outcomes into binary outcomes and then applying traditional MTL. For meta-analysis, we individually train models for each task and aggregate all models for subsequent prediction and biomarker identification. Four major machine-learning algorithms serve as base models in meta-analysis: lasso, ridge regression, random forest, and Support Vector Machine (SVM). For SVM, three kernels—"linear", "radial", and "polynomial"—are implemented."

## Real data analysis

**Prediction of Sepsis**

MTLComb is trained on clinical features to predict four outcomes: diagnosis (classification task) and the measurements (regression tasks) of lactate, urea, and creatinine. These regression tasks provide insights into the dynamics of the metabolic status and kidney function at sepsis onset. For comparison, we applied MTLBin and common machine learning (ML) methods as baselines. Details of cohorts and preprocessing steps are summarized in the supplementary methods.

Two cohorts are included in the analysis, where one serves as the training cohort and the other as the test cohort, and vice versa. This allows for cross-cohort prediction performance evaluation and biomarker reproducibility assessment. AUC is used as the metric for prediction performance. To quantify biomarker reproducibility, we compare the two models (one trained for one cohort) of each approach and count the number of overlapping features selected from the top 10 features. For MTLComb, only the classification model is used for testing. Comparison methods include MTLBin, Lasso[33], ridge regression[33], random forest[35], and SVM[36].

**Prediction of schizophrenia**

We aimed to identify aging-dependent genes associated with schizophrenia through two defined prediction tasks: predicting the diagnosis of schizophrenia (binary outcome) and predicting the age of subjects (continuous outcome). Given that MTLComb is designed specifically to capture age-dependent risk patterns, our expectation was not to achieve superior prediction performance compared to machine learning methods without such restrictions. Hence, this analysis does not focus on inter-method comparisons. Instead, the investigation revolves around whether MTLComb can capture gene markers predictive to all tasks and whether these markers can be validated in another cohort. Cohorts and preprocessing steps are detailed in the supplementary methods.

The analysis involves a discovery cohort and a validation cohort. In the discovery cohort, first, a 10-fold nested cross-validation procedure quantifies prediction performance. Here, regression and classification models are averaged to predict and identify shared markers. AUC is used for diagnosis prediction, and explained variance is calculated for age prediction. To account for sampling variability, the procedure is repeated 10 times, and the results are averaged. Next, the model trained on all subjects of the discovery cohort is validated on the separate validation cohort. Finally, to demonstrate the biological interpretability of our model, we re-trained the MTLComb model on both the discovery and validation cohorts, establishing four prediction tasks. The top 500 genes identified by the model were analyzed using the clusterProfiler[37] software for pathway enrichment analysis. Homogeneity of the selected genes is compared with other machine learning approaches.

# Results

## Simulation data analysis

In **Figure 2**, we present a performance comparison among different approaches using simulation data. **Figure 2** (a) illustrates the comparison of prediction performance. The superior explained variance of MTL approaches, in contrast to ML approaches, underscores the utility of joint modeling. As dimensionality increases, accurate prediction becomes challenging for all algorithms. MTLComb exhibits superior performance, followed by MTLBin. **Figure 2** (b) showcases the comparison of feature selection accuracy. As dimensionality increases, feature selection becomes more challenging for all algorithms. MTL approaches outperform meta-analysis methods. Among MTL approaches, MTLComb consistently demonstrates superior performance, especially in high-dimensional settings, highlighting its efficiency in utilizing data.

## Real data analysis

**Case study 1: Prediction of Sepsis**

The prediction results are presented in **Table 1**, where, in general, cohort 1 (training in cohort 2) exhibited better prediction performance than cohort 2 (training in cohort 1), potentially attributable to the more recent data from this cohort (cohort 2) with improved data quality. On average, MTLComb and Ridge regression achieved a similar prediction performance (average AUC $\approx$ 0.73). MTLComb outperformed Ridge on the prediction of cohort 1, while Ridge showed superior prediction performance in cohort 2. MTLBin was the second most accurate method.

In terms of model interpretability, we compared the model similarity and reproducibility of two representative methods—MTLComb and Ridge. **Table 2** illustrates that MTLComb models, trained on independent cohorts, exhibited higher similarity (r=0.7) compared to Ridge regression (r=0.41), highlighting the model stability against cross-cohort heterogeneity. On average, MTLComb identified four reproducible features, whereas Ridge regression yielded 1.2 (average over 10 repetitions), suggesting that MTLComb could pinpoint more reproducible features when applied to an unknown cohort.

Moreover, to demonstrate the biological relevance of MTLComb to the onset of sepsis, we tested the association between predicted scores and kidney function, as well as metabolic state measured on sepsis onset. As shown in **Table 2**, the MTLComb score demonstrated a higher association than Ridge, indicating that MTLComb could more accurately capture sepsis-relevant associations.

Finally, we list the four features and coefficients selected by MTLComb: 'SAPSII' (coefficient: 0.036), 'SOFA total score' (coefficient: 0.038), 'SIRS average λ'[38] (see **supplentary methods** for more detail, coefficient: 0.03), and 'SOFA cardiovascular score' (coefficient=0.029). The positive coefficients suggest that all these features were associated with an increment in sepsis risk. This is consistent with our expectation, as the SAPSII score summarizes 17 variables and was developed for the estimation of death risk on ICU admission[39]. The SOFA score represents a concurrent evaluation of organ dysfunction in 6 organ systems[40] and is an essential part of the current sepsis definition (sepsis-3[22]). A univariable association of the intensity of SIRS with sepsis risk for SIRS average λ has been shown in polytrauma patients[38]. With the exception of SOFA subscore for cardiovascular dysfuntion, the selected features all represent summary scores in which multiple parameters are assessed repeatedly, thereby likely representing patient state more efficiently than features assessed only once. Both SAPSII and SOFA are derived from the most extreme values (maximum or minimum) in the preceding 24 hours, and SIRS average λ reflects the intensity of systemic inflammation over 24 hours. Surprisingly both the total number of points (SOFA total score) as well as the score component reflecting the cardiovascular system were selected, underscoring the importance of circulatory failure for estimation of subsequent sepsis risk and concurrent renal function in polytrauma patients.

**Case study 2: Prediction of schizophrenia**

In the discovery data analysis, a 10-fold nested cross-validation indicates significant predictions for age (t= 2.833, p=0.005, r2=0.024) and diagnosis (t=3.1, p=0.0018, AUC=0.64) on unseen subjects in the discovery cohort. Subsequently, the model trained using all samples in the discovery cohort was tested on the validation cohort, yielding significant results for age prediction (t=3.06, p= 0.0029, r2= 0.084) and schizophrenia diagnosis association (t=4.1, p= 4.4e-05 , AUC=0.71).

Pathway analysis revealed 13 significantly enriched pathways, detailed in **Table S4**. Notably, several pathways exhibit strong associations with schizophrenia and aging. For instance, voltage-gated channel activity (GO:0022832, FDR=0.0031), chemical synaptic transmission (GO:0007268, FDR=0.026), trans-synaptic signaling (GO:0099537, FDR=0.026), and synaptic signaling (GO:0099536, FDR=0.034) have all been implicated in the development of schizophrenia and the aging process through their roles in synaptic plasticity[41,42].

To demonstrate that MTLComb allowed the selection of a more homogenous set of markers, the coefficients of the MTLComb model for both age- and diagnosis-prediction tasks are shown in **Figure 3** (a). For comparison, ridge and lasso models, trained individually for each outcome, are also plotted in **Figure 3** (c) and (b). Specifically, In **Figure 3** (b), individual lasso models select very different markers for each task. In **Figure 3** (c), the ridge regression models tend to consider the risk patterns of all features, individually predictive for each task. Therefore, due to complex correlations among genes in brain cohorts, many markers exhibited different behaviors in different tasks, complicating interpretation. In contrast, MTLComb identified the same markers with similarly predictive behaviors for all outcomes, as shown in **Figure 3** (a), which may index a shared molecular mechanism with higher likelihood.

## Discussion

In this manuscript, we introduce MTLComb, a novel MTL algorithm designed for the identification of shared biomarkers across mixed regression and classification tasks. At its core, MTLComb employs a provable losses weighting scheme, providing an analytical solution for optimal weights across different types of losses. Complementing this, we have developed an efficient optimization procedure, training protocol, and hyperparameter tuning procedure. MTLComb achieved state-of-the-art algorithmic complexity, making it applicable to large-scale problems. We demonstrated the efficacy of MTLComb through a simulation data analysis, showcasing its superior performance in prediction accuracy and joint feature selection, especially in high-dimensional settings. In the subsequent real data analysis of sepsis, MTLComb exhibited competitive prediction performance, increased model stability, higher marker selection reproducibility, and greater biological interpretability in the sepsis context. This positions MTLComb as a data integrative tool capable of aggregating subtle risk signals from heterogeneous clinical and biological datasets into a discernible and reproducible pattern. In the schizophrenia analysis, MTLComb successfully captured homogeneous gene markers predictive in both age- and diagnosis-prediction tasks, validated in an independent cohort. This suggests the potential of MTLComb to stratify age-dependent schizophrenia risk, offering novel applications in precision medicine. Here, we identified several synaptic signalling pathways that had previously been associated with both schizophrenia, as well as aging, likely due to their relevance for synaptic plasticity. For example, voltage-gated channel activity has been linked to schizophrenia due to its mediation of intracellular Ca2+ influx, which alters neuronal excitability and synaptic plasticity[43]. Interestingly, the regulation of Ca2+ also changes with aging, disrupting Ca2+ homeostatic processes and affecting the transmission of information across brain systems[44]. These transmission properties are crucial for the induction of synaptic plasticity. Our analysis supports the hypothesis that synaptic plasticity may represent a common pathway of relevance for both schizophrenia[45] and aging[46]. However, due to potential confounding factors such as medication effects in brain-expression, these findings should be interpreted with caution. The ability to identify signatures of shared relevance for different phenotys may make MTLComb useful for comorbidity analysis, as has been previously performed using regression-based MTL[47].

It is important to note that MTLComb has several limitations. With a regularization approach based on the linear model, MTLComb has favourable properties for analysis of high-dimensional problems but yields limited improvements in low-dimensional scenarios, as depicted in **Figure 2** (b). This observation is matched in the sepsis analysis presented in **Table 1**, where MTLComb demonstrated a prediction performance similar to that of ridge regression. Nevertheless, as indicated in **Table 2**, the utility of model interpretability and biological plausibility is still evident in the low-dimensional context despite the comparable prediction performance. This could be attributed to the inherent confounding in clinical data[48], where the confounding effect could be mitigated by learning information from multiple tasks using MTL. This aligns with our previous investigation[49] highlighting MTL's robustness against the cross-cohort and sampling variability when applied to heterogeneous cohorts. Second, although MTLComb greatly harmonized the feature selection principle of different types of tasks, there are still differences in the magnitude of the coefficients. An appropriate z-score standardization could minimize this problem, but further research is needed to solve this problem.

Future work may extend MTLComb by incorporating additional types of losses, broadening its application scope. For instance, in the sepsis analysis, adding a poisson regression model[50] to MTLComb could support the prediction of count data, such as length of ICU stay[26].

## Conclusion

In conclusion, MTLComb enhances the capabilities of linear MTL approaches by enabling learning from mixed types of tasks and facilitating unbiased joint feature selection through a provable losses weighting scheme. Its potential applications span various fields, such as comorbidity analysis and the simultaneous prediction of multiple clinical outcomes of diverse types.

## Code availability

https://github.com/transbioZI/MTLComb

## Acknowledgements

This study was supported through state funds approved by the State Parliament of Baden-Württemberg for the Innovation Campus Health + Life Science alliance Heidelberg Mannheim. The study was endorsed by the German Center for Mental Health (DZPG). This research was supported by the Intramural Research Program of the NIMH(NCT00001260, 900142).

# Figures

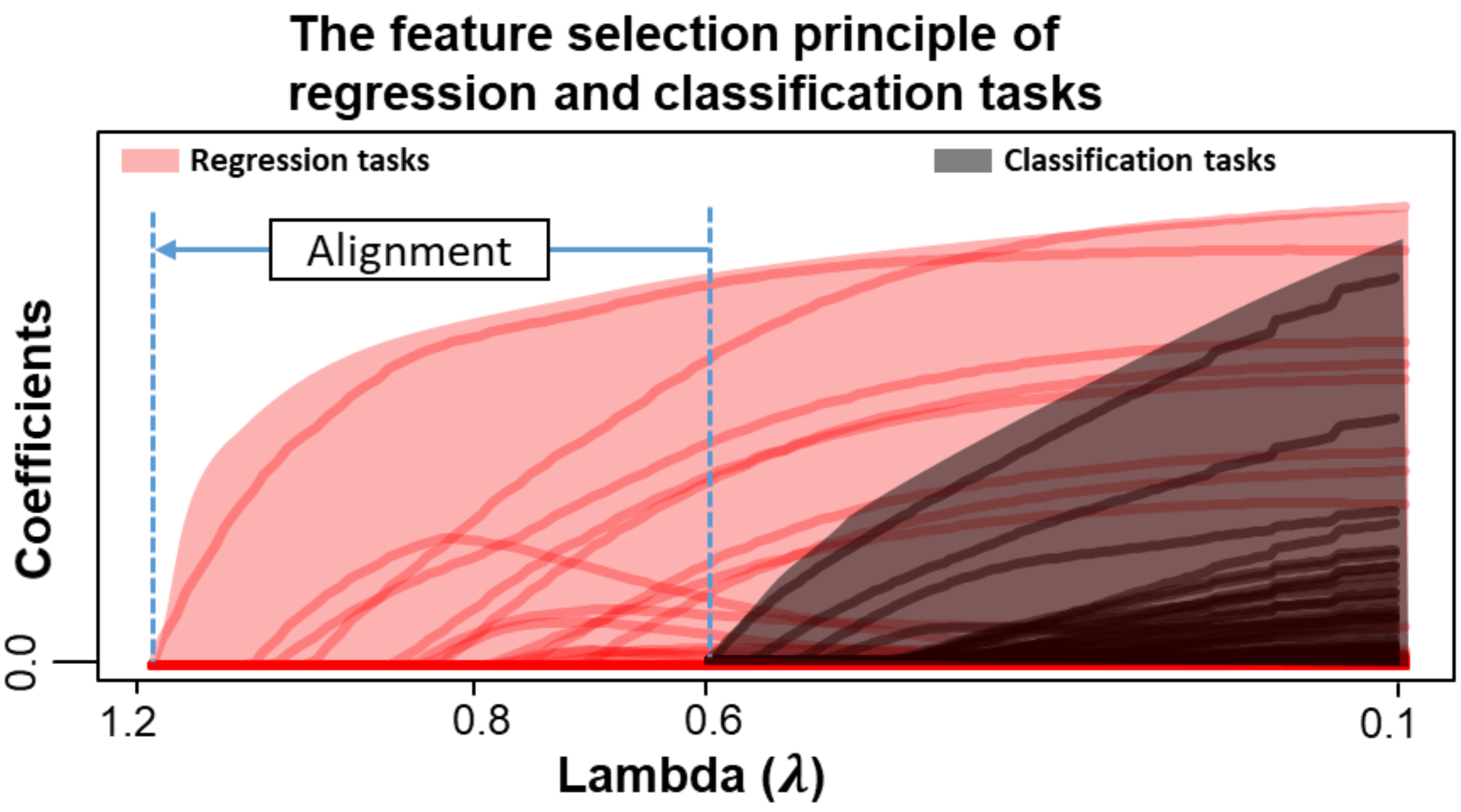

**Figure 1, Ilustration diagram of MTLComb.** The feature selection principle of linear MTL can be quantified as a regularization path[19]. A regularization path refered to a series of models with the continuous dyanmics of coefficients change, and is parameterized by a hyperparameter $\lambda$. A large value of $\lambda$ is associated with a fewer number of selected features. Here, the regularization path of regression/classification tasks are shown in red/black color. These pathes are not aligned due to the different type of losses, leading to a potential biased selection of joint features. For example, $\lambda = 0.8$ yields 7 features selected for regression tasks but null for classification. The MTLComb aimed to align the two regularization pathes.

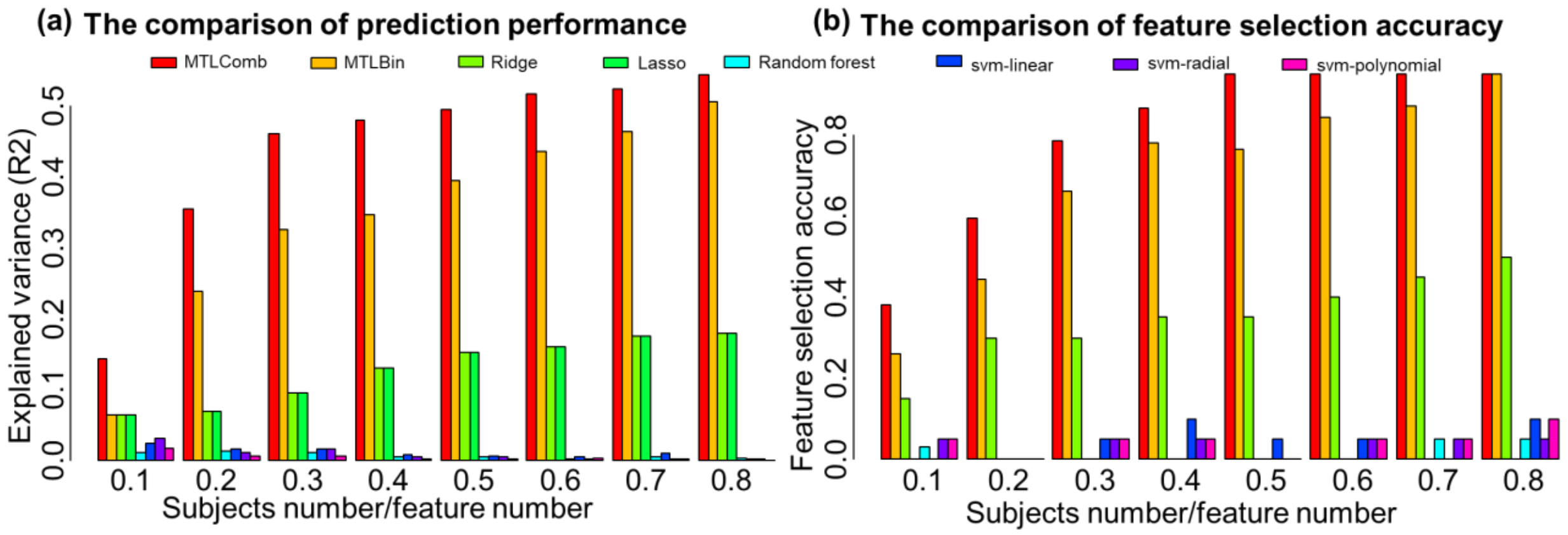

**Figure 2, Results of simulation data analysis.** (a), The comparison of prediction performance. The superior explained variance of MTL approaches compared to those of ML approaches demonstrate the utility of MTL. With increasing the dimensionality, accurate prediction becomes challenging for every algorithm. MTLComb outperforms other methods. (b) The comparison of joint feature selection accuracy. Increasing the dimensionality makes the feature selection more challenging for all algorithms. MTL approaches outperformed meta-analysis. Among MTL approaches, MTLComb outperformed MTLBin, particularly for high dimensional settings.

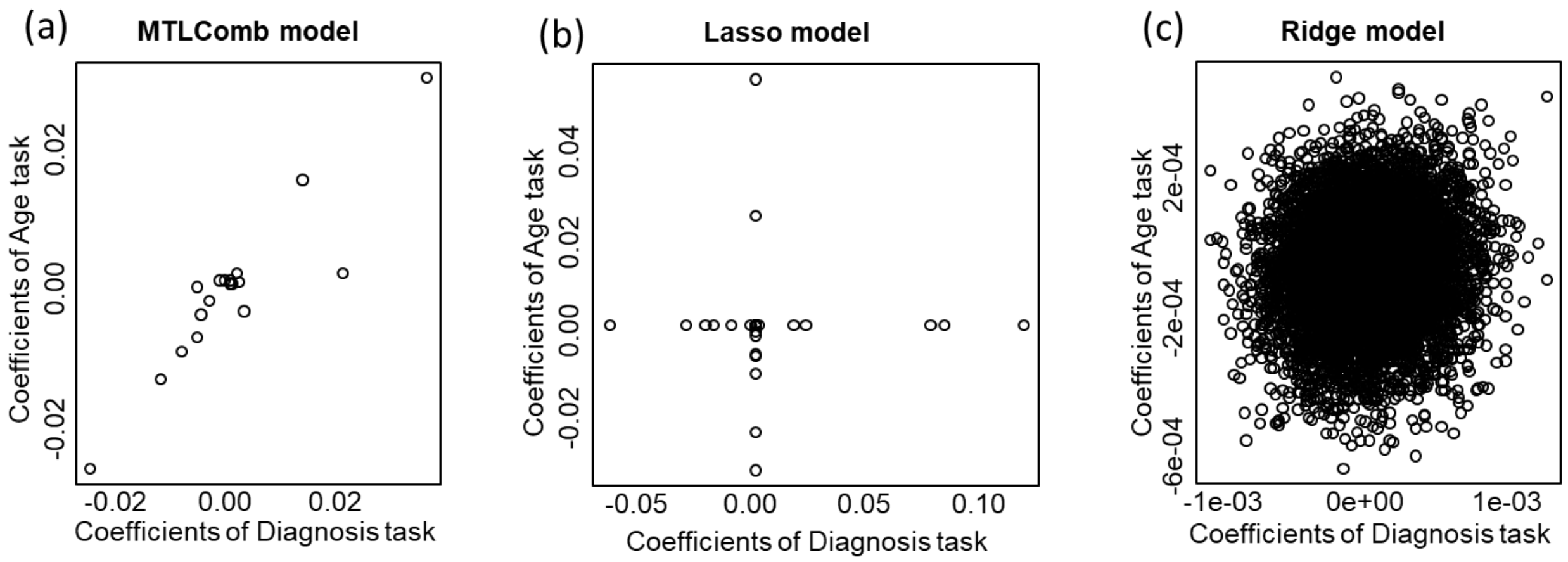

**Figure 3 Homogenous marker selection for age and diagnosis prediction.** These models are trained on the discovery cohort. MTLComb is trained simultaneously on two tasks. For Lasso and ridge regression, the model is trained for each outcome individually. The coefficients of both prediction tasks are shown for the MTLComb (a), Lasso (b) and Ridge (c) methods.

# Tables

| Metric | Test on cohort | MTLComb | MTLBin | Ridge | Lasso | Random forest | SVM-linear | SVM-radial | SVM-polynomial |
|---|---|---|---|---|---|---|---|---|---|
| AUC | 1 | **0.75** | 0.73 | 0.73 | 0.70 | 0.71 | 0.64 | 0.71 | 0.70 |
| | 2 | 0.71 | 0.70 | **0.73** | 0.68 | 0.67 | 0.65 | 0.72 | 0.69 |

Table 1, the cross-cohort prediction performance of early sepsis prediction.

| | MTLComb | Ridge regression |
|---|---|---|
| Model similarity | 0.70 | 0.41 |
| The number of reproducibly selected features | 4.0 | 1.2 |
| Predict lactate value (explained variance) | 10% | 4% |
| Predict on creatinine (explained variance) | 5.4% | 1.2% |
| Predict on urine (explained variance) | 15% | 5% |

Table 2, the model interpretability.

# MTLComb: multi-task learning combining regression and classification tasks for joint feature selection (Supplementary methods)

Han Cao[1], Sivanesan Rajan[2,3], Bianka Hahn[4], Ersoy Kocak[2,3], Daniel Durstewitz[1], Emanuel Schwarz[2,3]*, Verena Schneider-Lindner[4]*

1 Department of Theoretical Neuroscience, Central Institute of Mental Health, Medical Faculty, Heidelberg University, Germany

2 Department of Psychiatry and Psychotherapy, Central Institute of Mental Health, Medical Faculty, Heidelberg University, Mannheim, Germany

3 Hector Institute for Artificial Intelligence in Psychiatry, Central Institute of Mental Health, Medical Faculty Mannheim, Heidelberg University, Mannheim, Germany

4 Department of Anesthesiology and Surgical Intensive Care Medicine, Medical Faculty Mannheim, Heidelberg University, Theodor-Kutzer-Ufer 1-3, 68167, Mannheim, Germany

## Outline

## Modeling, optimization and algorithms

$$\min_{W} 2 \times \mathrm{Z}(W) + 0.5 \times \mathrm{R(W)} + \lambda||W||_{2,1} + \alpha||WG||_2^2 + \beta||W||_2^2 \quad (1)$$

where $\mathrm{Z}(W) = \sum_{i=1}^{c} \frac{1}{N_i} \log(1 + e^{-Y^{(i)}(X^{(i)}w^{(i)})})$ , $\mathrm{R}(W) = \sum_{i=c+1}^{t} \frac{1}{N_i} ||Y^{(i)} - X^{(i)}w^{(i)}||^2$ , $W = [w^{(1)}, \ldots w^{(c)}, \ldots w^{(t)}]$ and $G = \mathrm{diag}(t) - \frac{1}{t}$.

Among $t$ tasks, the first $c$ refers to the number of classification tasks, and the remaining to regression tasks. $\mathrm{Z}(W)$ is the logit loss to fit the classification model, and $\mathrm{R}(W)$is the least square loss to fit the regression model. $X = \left\{X^{(i)} \in \mathbb{R}^{N_i \times p} : i \in \{1, \ldots, c, \ldots t\}\right\}$, refers to the feature matrices of $t$ tasks, where $p$ features are consistent across tasks. $Y = \left\{Y^{(i)} \in \mathbb{R}^{N_i \times 1} : i \in \{1, \ldots, c, \ldots t\}\right\}$ describes the outcome lists associated with $c$ classification and $t - c$ regression tasks. $W \in \mathbb{R}^{p \times t}$ is the coefficient matrix that needs to be estimated, where each column represents the coefficient vector of each task, and each row $w^{(j)} \in \mathbb{R}^{t \times 1}$ consists of the coefficients of feature $j$. $||W||_{2,1} = \sum_{j=1}^{p} \sqrt{||w^{(j)}||_2^2}$ is a sparse penalty term to promote the joint feature selection[1]. $||WG||_2^2$ is the mean-regularized term[2,3] to promote the selection of features with similar cross-task coefficients. $||W||_2^2$ aims to select the correlated features and stabilize the numerical sulutions[4]. $\{\lambda, \alpha, \beta\}$ is the hypermeter set, which controls the strength of the penalty. Here, $\lambda$ is selected by cross-validation, and users select the rest as constant prior. We weight $\mathrm{Z}(W)$ by 2 and $\mathrm{R}(W)$ by 0.5. This simple weighting scheme made the regularization paths consistent.

The MTLComb for integrating regression and classification tasks with joint feature selection is formulated as (1). This problem is not easy to solve due to the non-smooth penalty. To solve it efficiently, we adopt the accelerated proximal gradient descent method to approximate the solution, yielding state-of-art efficiency.

Let $F(W) = 2 \times \mathrm{Z}(W) + 0.5 \times \mathrm{R(W)} + \alpha||WG||_2^2 + \beta||W||_2^2$ and $\Omega(W) = ||W||_{2,1}$, the formulation (1) is re-written as (2),

$$\min_{W} F(W) + \lambda\Omega(W) \quad (2)$$

$F(W)$ is a smooth and convex function about $W$, and $\Omega(W)$ is the non-smooth and convex. Assume the Lipschitz constant of $F(W)$ is $L$, the key step is to solve the following subproblem (3) in each iteration,

$$W_{i+1} = \arg\min_{y} F(W_i) + \langle \nabla F(W_i), \mathrm{y} - W_i \rangle + \frac{L}{2} \left|\left| y - W_i \right|\right|_2^2 + \lambda\Omega(y) \tag{3}$$

Problem (3) is the second-order approximation of $F(.)$ given the current standing point $W_i$. After re-organization, we have an equivalent form (4),

$$W_{i+1} = \arg\min_{y} \frac{L}{2} \left( y - \left( W_i - \frac{\nabla F(W_i)}{L} \right) \right)^2 + \lambda ||y||_{2,1} \tag{4}$$

Let $w^{(j)}{}_i \in \mathbb{R}^{1 \times t}$ is the $j$th row of $W_i$, problem (4) can be divided into $p$ independent small problems to solve. Such a "divide and conquer" strategy greatly reduces the complexity of the problem. Each small problem took the form (5),

$$w^{(j)}{}_{i+1} = \arg\min_{y} \frac{L}{2} \left( y - \widetilde{w}^{(j)}{}_i \right)^2 + \lambda \sqrt{\left|\left| y^{(j)} \right|\right|^2} \tag{5}$$

where

$$\widetilde{w}^{(j)}{}_i = w^{(j)}{}_i - \frac{\nabla F(w_i)^{(j)}}{L}$$

This problem (5) can be solved analytically

$$w^{(j)}{}_{i+1} = \left( 1 - \frac{\lambda/_L}{\max\left\{ \left|\left| \widetilde{w}^{(j)}{}_i \right|\right|_2, \lambda/_L \right\}} \right) \widetilde{w}^{(j)}{}_i \tag{6}$$

Since the Lipschitz constant L is usually unknown, we estimated it using the ajimo-goldstein line search[5] in MTLComb. Since the objective of MTLComb is convex and non-smooth, we accelerate the entire procedure with the momentum methods. Here, we applied Nesterov's acceleration approach [6-8]. For each iteration $i$, a search point was quantified by weighing the estimates from the previous two iterations,

$$\mathrm{s_i} = \frac{\alpha_{i-1}}{\alpha_i} \mathrm{w_i} + \frac{1 - \alpha_{i-1}}{\alpha_i} \mathrm{w_{i-1}} \tag{7}$$

The search point was subsequently sent to (6) to update the solution. The procedure will repeat until the solution converges to the required precision. The entire algorithm integrating all these techniques is summarized in **Algorithm 1.**

| **Algorithm 1** The solver of MTLComb |
| --- |
| **Input**: $\lambda > 0, L_0 > 0, W_0, maxIter > 0$<br>**Output**: $W_{i+1}$<br>1: Initialize $W_1 = W_0$, $\alpha_{-1} = \alpha_0 = 0$, and $L = L_0$<br>2: **for** $i = 1$ to $maxIter$ **do**<br>3: Set $\mathrm{S_i} = \mathrm{W_i} + \frac{\alpha_{i-1}-1}{\alpha_i}(\mathrm{W_i} - \mathrm{W_{i-1}})$<br>4: Find the smallest $L \in \{L_{i-1}, 2L_{i-1}, 4L_{i-1}, 16L_{i-1}, \dots\}$ such that<br>$F(\mathrm{S_i}) + \langle \nabla F(\mathrm{S_i}), W_{i+1} - \mathrm{S_i}\rangle + \frac{L}{2}\|W_{i+1} - W_i\|_2^2 \geq F(W_{i+1}),$<br>where $W_{i+1}$ is quantified according to (4) and (5)<br>5: Set $L_i = L$, and $\alpha_{i+1} = \frac{1+\sqrt{1+4\alpha_i^2}}{2}$<br>6: If the termination rule is satisfied, **return**<br>7: **end for** |

## Proof of proposition 1

**Proposition 1.** *By weighting* $\mathrm{Z}(W)$ *with 2 and* $\mathrm{R(W)}$ *with 0.5, we quantify an identical form of lam_max for both the regression and the classification tasks*

**Proof:**

Let m be the index of a classification task, j be the index of a feature. Then $w_{jm}$ is an element of the coefficient matrix $W_c$ of classification tasks. $w_{j,}$ refers to the jth column of $W_c$, representing the coefficients of feature j. $w_{,m}$ refers to the ith row of $W_c$, representing the coefficients of task m. $N_m$ refers to the number of subjects of task m.

The subgradient about the coefficient $w_{jm}$ took the form (8),

$$\partial_{w_{jm}} = \frac{1}{N_m}\sum_{i=1}^{N_m}\frac{-y_j^{(m)}x_{ij}^{(m)}}{1+\exp(y_j^{(m)}x_i^{(m)}w_{,m})} + 2\beta w_{jm} + 2\alpha GGw_{j,}{}^T + \lambda v_j, \tag{8}$$

$$v = \{x \in R^t : \|x\|_2 \leq 1\}$$

Let $\partial_{w_{jm}} = 0$ and $W_c = \mathbf{0}$, we have

$$\partial_{w_{jm}} = -\frac{1}{N_m}\sum_{i=1}^{N_m} y_j^{(m)} x_{ij}^{(m)} + \lambda v_j = 0$$

$$\lambda = \frac{\sum_{i=1}^{N_m} y_j^{(m)} x_{ij}^{(m)}}{N_m v_j}, \text{ subject to } v = \{x \in R^t : ||x||_2 \leq 1\} \quad (9)$$

The smallest $\lambda$ that satisfies this equation is

$$\lambda^* = \min_{v_j} \frac{\sum_{i=1}^{N_m} y_j^{(m)} x_{ij}^{(m)}}{N_m v_j} = \frac{\sum_{i=1}^{N_m} y_j^{(m)} x_{ij}^{(m)}}{N_m} \quad (10)$$

Therefore, $\lambda^{*(m)} = \frac{\sum_{i=1}^{N_m} y_j^{(m)} x_{ij}^{(m)}}{N_m}$ for the classification task m

Similarly, for a regression task n, we assume the coefficient $w_{jn}$ an element of the coefficient matrix $W_r$ of regression tasks. $w_{j,}$ refers to the jth column of $W_r$, representing the coefficients of feature j. $w_{,n}$ refers to the ith row of $W_r$, representing the coefficients of task n. $N_n$ refers to the number of subjects of task n.

The subgradient about the coefficient $w_{jn}$ is

$$\partial_{w_{jn}} = \frac{2}{n_n} X^{(n)^T}(X^{(n)} w_{,n} - y^{(n)}) + 2\beta w_{jn} + 2\alpha G G w_{j,}{}^T + \lambda v_j, \quad (11)$$
$$v = \{x \in R^t : ||x||_2 \leq 1\}$$

After applying the same procedure as for the classification task, we give directly the solution for regression task n, $\lambda^{*(n)} = \frac{\sum_{i=1}^{N_n} y_j^{(n)} x_{ij}^{(n)}}{N_n}$

It is clear that $\lambda^{*(m)}$ has the same form as $\lambda^{*(n)}$

**The proof is completed.**

The entire algorithm for estimating the regularization path is shown in **Algorithm 2**

| **Algorithm 2** The estimation of the regularization path |
| --- |
| **Input**: $\lambda_1 > \lambda_2 > \cdots > 0$ |
| **Output**: $W_1, W_2, \ldots$ |
| 1: Initialize $W_0 = p \times t = 0$ |
| 2: **for** $i = \{1, 2, \ldots\}$ **do** |
| 3: $W_i =$ **Algorithm 1** ($\lambda = \lambda_i, L_0 = 1, W_0 = W_{i-1}, maxIter = 100$) |
| 4: **end for** |

**Parameter tuning**

The optimal $\lambda$ is determined by k-fold cross-validation (CV) in MTLComb. $\alpha$ and $\beta$ are determined by the users as prior. The intercept and intercept-free models are both provided to fit normalized and unnormalized data, respectively.

## Data simulation protocol

10 classification and 10 regression tasks were simulated. Let the task number t=10, feature number p=200, and sample size N=100. The coefficient matrix is sampled from normal distribution $\mathrm{W} \sim \aleph_{p\times t}(0,1)$. Then, 90% of features are set to 0, creating sparsity $\mathrm{W}[,(0.1\mathrm{p}+1){:}\,\mathrm{p}] = 0$. For the task m, the feature matrix is created by sampling from the normal distribution $X^{(m)} \sim \aleph_{N_m\times p}(0,1)$.

**Regression task**

To obtain the outcome, we first multiplied the feature matrix and coefficient vector and then added the random noise, i.e., $Y^{(m)} = X^{(m)} \times \mathrm{W}[,\mathrm{m}] + 0.5\aleph_{N_m\times 1}(0,1)$.

**Classification task**

To obtain the outcome, we first multiplied the feature matrix and coefficient vector, then added the random noise, and subsequently binarized the resulting scores, i.e.

$$Y^{(m)} = \left[X^{(m)} \times \mathrm{W}[,\mathrm{m}] + 0.5\aleph_{N_m\times 1}(0,1)\right]_+$$

Here, the strength of noise is set to 50% of the data signal. Using the same technique, we created 20 additional tasks as the test data.

## Case study 1: Prediction of sepsis

**Data cohorts**

Two cohorts of polytrauma patients were retrospectively created from the electronic medical records of the surgical ICU of the University Medical Center Mannheim, Germany. Cohort 1 consisted of 242 patients treated between 2006 and 2011[9], and containeds 78 sepsis cases and 164 controls with sufficient data for this study. Cohort 2 with 159 patients was treated in the ICU thereafter, from 2011 to 2016 and contained 56 sepsis cases and 103 controls.

**Data preprocessing and outcome determination**

The first measurement up to 24 hours after admission of 57 numerical features representing routine laboratory values, vital signs and interventions as well as clinical scores were included as potential predictors in the analysis (**Table S3**). These also included the criteria of the systemic inflammatory response syndrome (SIRS), which are derived from leucocyte count, heart rate, temperature and respiratory rate[10]. The SIRS-criteria were represented by descriptors which summarize SIRS over the 24 hours after admission as previously described[9]: in each patient, an algorithm determined the number of SIRS criteria met in each minute (λ). From this, the mean of 1440 minute λs was calculated for determination of 'SIRS average λ', the difference between λ of the last and the first minute of the 24 hours for 'SIRS Δ' and the number of changes in λ during the 24 hour period as 'SIRS C'. Multiple-imputation was applied to impute missing feature values. MCMC was used as the base model in the imputation, and 100 different imputed data tables were outputted and then averaged for the subsequent process. The imputation was performed using the internal functions of the software SAS V 9.4. Sepsis incidence was determined by retrospective review of the clinical data by medical experts. For the septic patients, the continuous coutcomes were determined as the last urea, creatinine and lactate measurements up to 24 hours before the sepsis onset time point. To determine the continuous outcome values of the non-septic polytrauma patients, we first created risk sets[11] at each time point a sepsis case was diagnosed during follow-up time in the ICU. The risk set contained the case and all non-cases with the same or longer duration of treatment in the ICU with respect to this time point. For each patient remaining sepsis-free until ICU discharge we took the meadian of all risks sets in which the individual patient had been included. If the median was between 2 risk sets, the earlier risk set, i.e. the one corresponding to the case with shorter time between admission and sepsis onset was chosen. The values that represented the last urea, creatinine and lactate measurements up to 24 hours before this time point were analyzed as continuous outcome parameters.

Before the machine learning stage, we performed a z-standardization on the feature matrix and the continuous outcomes.

## Case study 2: Prediction of schizophrenia

**Data cohorts**

Four indepdent cortical expression cohorts were included in this study. The discovery dataset used for algorithm training was from the HBCC (Human Brain Collection Core) (n=422) cohort comprising genome-wide gene expression data quantified by microarray (dbGAP ID: phs000979.v3.p2). All patients included in the cohort met the Diagnostic and Statistical Manual of Mental Disorders, Fourth Edition (DSM-IV) criteria for a lifetime diagnosis of Axis I psychiatric disorders, such as schizophrenia or schizoaffective disorder, bipolar disorder, and major depression. A detailed description of this dataset can be found in the original publication[12]. The validation dataset, comprising 194 subjects, integrates three cohorts: GSE53987, GSE21138, and GSE35977. These cohorts are publicly available for download from the Gene Expression Omnibus (GEO) repository. Detailed descriptions of these cohorts can be found on the GEO platform and in their respective original studies[13-15].

**Data Preprocessing**

The **discovery dataset** was normalised and quality controlled using the R package lumi 2.48.0[16]. First, we extracted Illumina BeadArray data from IDAT for HumanHT-12 v4 Gene Expression BeadChip based on the binary manifest file using the function LumiR.idat(). Then we selected samples from the DLPFC region, individuals aged 18-65 years, of either African American (AA) or Caucasian (CAUC) ethnicity. Here, a different procedure was applied to diagnosis-prediction and age-prediction tasks. For the diagnosis-prediction task, the data from all healthy controls and schizophrenia patients were used for the subsequent analysis, using age, age², race, RIN, sex, pH, PMI, 5 principle components and 5 surogate variables as covariates. For the age-prediction task, only healthy controls were investigated and ethnicity, RIN, sex, pH, PMI, 5 principle components and 5 surogate variable used as covariates. The following steps are applied individually for each task.

Next, we corrected for background noise using the lumiExpresso() function with quantile normalisation and log2 transformation parameters. We retained robustly expressed probes as those with a detection p-value <0.01 in at least half of the individuals. We then excluded features with unknown expression (unavailable) and eliminated duplicate features. In order to eliminate duplicate features, we ordered them in descending order based on their median expression values across samples, selecting the highest value and discarding the others. Subsequently, the Ensembl IDs associated with the features were converted to gene symbols using the ensembl-based annotation R package (lumiHumanAll.db v1.22.0). Outlier samples deviating more than 4 standard deviations from the first or second PC were then removed. Prior to Surrogate Variable Analysis (SVA), missing values for the covariates pH and PMI were imputed using the mean of the available data. SVA was performed with control for the effects of the covariates. Then, PCs were calculated and the data adjusted for the top 5 PCs, the top 5 surrogate variables and the covariates. Finally, the resulting expression genes were z-standardized. The cohort comprised data for 201 healthy controls and 158 patients with schizophrenia that was used for the diagnosis-prediction task, and for 283 healthy controls for the age-prediction task. The demongraphics of these subjects are shown in **Table S1**. In the end, 9663 genes are included in the discovery cohort for in-cohort prediction test.

The **validation dataset** consists of three cohorts: GSE53987, GSE21138, and GSE35977. Initially, raw data were extracted using the ReadAffy() function from the R package affy[17]. This was followed by normalization via the Robust Multi-array Average (RMA) method[18]. Subsequently, probe IDs were converted into gene symbols for each cohort, tailored to the specific type of microarray chip used. For cohorts GSE53987 and GSE21138, the hgu133plus2.db database was employed, corresponding to data acquired on the Affymetrix GeneChip Human Genome U133 Plus 2.0 Array. For the GSE35977 cohort, which utilized the Affymetrix Human Gene 1.0 ST Array, the hugene10sttranscriptcluster.db database was used. Data alignment across the three cohorts involved mapping gene names and averaging values from multiple probes associated with the same gene. Subjects younger than 18 or older than 65 were excluded from the analysis. Distinct methodologies were then applied based on the prediction task: diagnosis-prediction task utilized data from both healthy controls and schizophrenia patients, incorporating covariates such as age, age squared, sex, pH, post-mortem interval (PMI), cohort index, five principal components, and five surrogate variables. Conversely, the age-prediction task was confined to healthy controls using the covariates: sex, pH, post-mortem interval (PMI), cohort index, five principal components, and five surrogate variables. The following steps are applied individually for each prediction task.

Outliers within each cohort were removed, defined as those exceeding four standard deviations from the mean along the first two principal components. Then, the cohorts were merged for surrogate variable analysis (SVA) using five surrogate variables. The top five principal components (PCs) were calculated for correction purposes. Then a linear regression analysis was utilized to adjust for potential confounders, incorporating all covariates, five surrogate variables, and five PCs. Finally, the gene expression data were z-standardized. This process resulted in a dataset comprising 92 healthy controls and 93 schizophrenia patients for the diagnosis-prediction task, and 92 healthy controls for the age-prediction task. The demographics of these subjects are detailed in **Table S2**. Currently, 17,166 genes are under consideration. After matching with genes from the discovery cohort, 8,799 genes were retained in the validation cohort for cross-cohort prediction testing.

| Parameter/Summary | Healthy controls | | SCZ | | Total | |
|---|---|---|---|---|---|---|
| | *n* | % | *n* | % | *n* | % |
| Total | 201 | 56.0 | 158 | 44 | 359 | 100 |
| Gender | | | | | | |
| Female | 58 | 28.9 | 55 | 34.8 | 113 | 31.5 |
| Male | 143 | 71.1 | 103 | 65.2 | 246 | 68.5 |
| Race | | | | | | |
| AA | 108 | 53.7 | 65 | 41.1 | 173 | 48.2 |
| CAUC | 93 | 46.3 | 93 | 58.9 | 186 | 51.8 |
| Age | *year* | | *year* | | *year* | |
| mean | 43.0 | | 45.9 | | 44.3 | |
| median | 45.4 | | 47.2 | | 46.2 | |

| | | | |
|---|---|---|---|
| min/max | 18.0 / 65.0 | 18.0 / 63.2 | 18.0 / 65.0 |
| 1st quantile | 31.6 | 38.8 | 35.6 |
| 3rd quantile | 53.6 | 54.4 | 54.0 |

**Table S1:** Demongraphics of the discovery dataset (HBCC)

| Parameter/Summary | Healthy controls | | SCZ | | Total | |
|---|---|---|---|---|---|---|
| | *n* | % | *n* | % | *n* | % |
| Total | 92 | 50 | 93 | 50 | 185 | 100 |
| Gender | | | | | | |
| Female | 26 | 28 | 27 | 29.0 | 53 | 28.6 |
| Male | 66 | 72 | 66 | 71.0 | 132 | 71.4 |
| Age | *year* | | *year* | | *year* | |
| mean | 44.1 | | 42.5 | | 43.3 | |
| median | 45 | | 45 | | 45 | |
| min/max | 21.0 / 65.0 | | 19.0 / 65.0 | | 19.0 / 65.0 | |
| 1st quantile | 37.8 | | 35.0 | | 36.0 | |
| 3rd quantile | 50.0 | | 50.0 | | 50.0 | |

**Table S2:** Demongraphics of the validation dataset (GEO cohorts)

| | Cohort 1 (N=242) | | | Cohort 2 (N=159) | | |
|---|---|---|---|---|---|---|
| | Sepsis (N=78) | No sepsis (N=164) | | Sepsis (N=56) | No sepsis (N=103) | |
| | Mean (SD) Median (IQR) n (%) | Mean (SD) Median (IQR) n (%) | p-value | Mean (SD) Median (IQR) n (%) | Mean (SD) Median (IQR) n (%) | p-value |
| Age [yrs] | 50.8 (20.82) | 47.0 (19.26) | 0.1777 | 52.6 (18.88) | 52.1 (20.96) | 0.8912 |
| Male | 64 (82.1%) | 120 (73.2%) | 0.1304 | 47 (83.9%) | 67 (65.0%) | 0.0116 |
| SAPSII* | 29 (25–37) | 25 (19–31) | <.0001# | 32 (25–38) | 29.5 (23–36) | 0.0354# |
| ISS | 41 (34–41) | 34 (29–41) | <.0001# | 29 (24–36) | 26 (22–34) | 0.0266# |
| AIS Head | | | 0.5815~ | | | 0.5361~ |
| 0 | 33 (42.3%) | 74 (45.1%) | | 14 (25.0%) | 33 (32.0%) | |
| 1 | 0 (0%) | 1 (0.61%) | | 0 (0%) | 1 (0.97%) | |
| 2 | 3 (3.85%) | 9 (5.49%) | | 3 (5.36%) | 1 (0.97%) | |
| 3 | 7 (8.97%) | 24 (14.6%) | | 5 (8.93%) | 7 (6.80%) | |
| 4 | 34 (43.6%) | 54 (32.9%) | | 31 (55.4%) | 54 (52.4%) | |
| 5 | 1 (1.28%) | 2 (1.22%) | | 3 (5.36%) | 7 (6.80%) | |
| AIS Face | | | 0.9898~ | | | 0.5292~ |
| 0 | 40 (51.3%) | 87 (53.0%) | | 30 (53.6%) | 66 (64.1%) | |
| 1 | 1 (1.28%) | 2 (1.22%) | | 6 (10.7%) | 9 (8.74%) | |
| 2 | 10 (12.8%) | 17 (10.4%) | | 19 (33.9%) | 27 (26.2%) | |
| 3 | 23 (29.5%) | 48 (29.3%) | | 1 (1.79%) | 1 (0.97%) | |
| 4 | 4 (5.13%) | 9 (5.49%) | | 0 (0%) | 0 (0%) | |
| 5 | 0 (0%) | 1 (0.61%) | | 0 (0%) | 0 (0%) | |
| AIS Thorax | | | 0.0113~ | | | 0.1191~ |
| 0 | 11 (14.1%) | 21 (12.8%) | | 11 (19.6%) | 30 (29.1%) | |
| 1 | 0 (0%) | 1 (0.61%) | | 2 (3.57%) | 6 (5.83%) | |

| | Cohort 1 (N=242) | | | Cohort 2 (N=159) | | |
|---|---|---|---|---|---|---|
| | Sepsis (N=78) | No sepsis (N=164) | | Sepsis (N=56) | No sepsis (N=103) | |
| | Mean (SD) Median (IQR) n (%) | Mean (SD) Median (IQR) n (%) | p-value | Mean (SD) Median (IQR) n (%) | Mean (SD) Median (IQR) n (%) | p-value |
| 2 | 1 (1.28%) | 7 (4.27%) | | 5 (8.93%) | 6 (5.83%) | |
| 3 | 17 (21.8%) | 66 (40.2%) | | 16 (28.6%) | 38 (36.9%) | |
| 4 | 49 (62.8%) | 69 (42.1%) | | 20 (35.7%) | 23 (22.3%) | |
| 5 | 0 (0%) | 0 (0%) | | 2 (3.57%) | 0 (0%) | |
| AIS Abdomen | | | 0.1300 | | | 0.1828~ |
| 0 | 37 (47.4%) | 91 (55.5%) | | 28 (50.0%) | 69 (67.0%) | |
| 1 | 0 (0%) | 0 (0%) | | 1 (1.79%) | 0 (0%) | |
| 2 | 3 (3.85%) | 13 (7.93%) | | 14 (25.0%) | 18 (17.5%) | |
| 3 | 13 (16.7%) | 30 (18.3%) | | 6 (10.7%) | 7 (6.80%) | |
| 4 | 23 (29.5%) | 29 (17.7%) | | 7 (12.5%) | 9 (8.74%) | |
| 5 | 2 (2.56%) | 1 (0.61%) | | | | |
| AIS Extremities | | | 0.8554~ | | | 0.7922~ |
| 0 | 14 (17.9%) | 38 (23.2%) | | 16 (28.6%) | 27 (26.2%) | |
| 1 | 0 (0%) | 1 (0.61%) | | 1 (1.79%) | 3 (2.91%) | |
| 2 | 4 (5.13%) | 10 (6.10%) | | 15 (26.8%) | 35 (34.0%) | |
| 3 | 32 (41.0%) | 63 (38.4%) | | 12 (21.4%) | 23 (22.3%) | |
| 4 | 28 (35.9%) | 52 (31.7%) | | 12 (21.4%) | 14 (13.6%) | |
| 5 | | | | | 1 (0.97%) | |
| AIS Soft tissue | | | 0.1225 | | | 0.8869~ |
| 0 | 5 (6.41%) | 15 (9.15%) | | 0 (0%) | 1 (0.97%) | |
| 1 | 0 (0%) | 9 (5.49%) | | 20 (35.7%) | 41 (39.8%) | |
| 2 | 45 (57.7%) | 98 (59.8%) | | 35 (62.5%) | 59 (57.3%) | |
| 3 | 25 (32.1%) | 35 (21.3%) | | 1 (1.79%) | 2 (1.94%) | |
| 4 | 3 (3.85%) | 7 (4.27%) | | 0 (0%) | 0 (0%) | |
| Diabetes | 10 (12.8%) | 6 (3.66%) | 0.0073 | 7 (12.5%) | 13 (12.6%) | 0.9824 |
| Respiratory diseases | 4 (5.13%) | 6 (3.66%) | 0.7310~ | 3 (5.36%) | 1 (0.97%) | 0.1257~ |
| Alcoholism | 14 (17.9%) | 17 (10.4%) | 0.0990 | 12 (21.4%) | 10 (9.71%) | 0.0409 |
| Cardiovascular diseases | 15 (19.2%) | 17 (10.4%) | 0.0571 | 17 (30.4%) | 9 (8.74%) | 0.0004 |
| Body temperature [°C] | 35.7 (1.45) | 36.2 (1.25) | 0.0094 | 36.1 (1.05) | 36.0 (1.35) | 0.5294 |
| pH* | 7.338 (0.08) | 7.351 (0.07) | 0.2035 | 7.334 (0.10) | 7.369 (0.06) | 0.0169 |
| Lactate [mmol/L]* | 2.34 (1.74) | 1.90 (1.13) | 0.0403 | 2.16 (1.51) | 1.91 (1.19) | 0.2861 |
| Base excess [mmol/L]* | -1.5 (3.43) | -1.0 (2.77) | 0.3207 | -2.1 (3.09) | -1.1 (2.47) | 0.0427 |

| | Cohort 1 (N=242) | | | Cohort 2 (N=159) | | |
|---|---|---|---|---|---|---|
| | Sepsis (N=78) | No sepsis (N=164) | | Sepsis (N=56) | No sepsis (N=103) | |
| | Mean (SD) Median (IQR) n (%) | Mean (SD) Median (IQR) n (%) | p-value | Mean (SD) Median (IQR) n (%) | Mean (SD) Median (IQR) n (%) | p-value |
| pO2 [mmHg]* | 177.5 (109.18) | 160.2 (89.43) | 0.2274 | 171.9 (105.44) | 151.8 (82.11) | 0.2226 |
| pCO2 [mmHg]* | 45.2 (10.31) | 44.6 (8.32) | 0.6897 | 47.6 (21.21) | 42.3 (7.28) | 0.0738 |
| Calcium [mmol/L]* | 1.1 (0.11) | 1.1 (0.08) | 0.2192 | 1.2 (0.08) | 1.2 (0.09) | 0.3549 |
| Potassium [mmol/L]* | 3.9 (0.59) | 3.9 (0.46) | 0.7909 | 4.0 (0.54) | 4.0 (0.44) | 0.4369 |
| FiO2 [%] | 55.3 (22.59) | 47.1 (23.97) | 0.0104 | 47.5 (19.35) | 38.4 (17.43) | 0.0040 |
| Mechanical Ventilation | 67 (85.9%) | 109 (66.5%) | 0.0015 | 48 (85.7%) | 70 (68.0%) | 0.0145 |
| WBC [10E9/L]* | 9.08 (3.96) | 10.56 (4.76) | 0.0121 | 11.98 (4.98) | 10.69 (3.71) | 0.0954 |
| Hb [g/dL] | 9.65 (2.39) | 10.93 (2.20) | 0.0001 | 10.71 (2.14) | 11.02 (2.08) | 0.3908 |
| Hematocrit [%]* | 27.72 (6.63) | 31.06 (5.41) | 0.0002 | 28.54 (6.53) | 30.85 (5.72) | 0.0284 |
| Erythrocytes [10E12/L]* | 3.2 (0.75) | 3.6 (0.62) | 0.0002 | 3.3 (0.72) | 3.5 (0.65) | 0.0171 |
| Thrombocytes [10E9/L]* | 120.7 (47.24) | 160.6 (63.05) | <.0001 | 158.3 (63.22) | 168.6 (52.27) | 0.3027 |
| INR* | 1.2 (0.29) | 1.1 (0.19) | 0.0290 | 1.2 (0.40) | 1.1 (0.17) | 0.1544 |
| pTT [sec]* | 30.2 (8.20) | 26.8 (4.00) | 0.0006 | 26.7 (4.60) | 26.0 (4.32) | 0.3539 |
| Mean arterial pressure (MAP) [mmHg] | 88.2 (18.26) | 89.9 (16.41) | 0.4928 | 84.4 (17.36) | 86.6 (18.75) | 0.4538 |
| Systolic blood pressure [mmHg] | 125.4 (27.36) | 130.1 (22.23) | 0.1858 | 122.3 (27.38) | 128.2 (32.44) | 0.2217 |
| Diastolic blood pressure [mmHg] | 68.72 (14.54) | 70.95 (13.88) | 0.2604 | 66.55 (13.97) | 68.76 (16.08) | 0.3699 |
| Heart rate [1/min] | 91.6 (21.21) | 88.9 (18.87) | 0.3275 | 90.8 (21.79) | 86.9 (23.50) | 0.2862 |
| Catecholamines | 34 (43.6%) | 33 (20.1%) | 0.0001 | 44 (78.6%) | 45 (43.7%) | <.0001 |
| Shock index | 0.78 (0.28) | 0.70 (0.20) | 0.0489 | 0.78 (0.27) | 0.71 (0.25) | 0.1266 |
| Volume balance [L]* | 4.25 (2.70) | 3.64 (2.73) | 0.1184 | 4.09 (3.30) | 3.03 (2.83) | 0.0444 |
| Bilirubin [mg/dL] | 0.76 (0.45) | 0.65 (0.38) | 0.0851 | 0.76 (0.50) | 0.72 (0.45) | 0.6743 |

| | Cohort 1 (N=242) | | | Cohort 2 (N=159) | | |
|---|---|---|---|---|---|---|
| | Sepsis (N=78) | No sepsis (N=164) | | Sepsis (N=56) | No sepsis (N=103) | |
| | Mean (SD) Median (IQR) n (%) | Mean (SD) Median (IQR) n (%) | p-value | Mean (SD) Median (IQR) n (%) | Mean (SD) Median (IQR) n (%) | p-value |
| Glucose [mg/dL]* | 146.0 (45.10) | 136.1 (36.25) | 0.0930 | 146.9 (40.84) | 140.8 (41.99) | 0.3738 |
| Creatinine [mg/dL]* | 0.96 (0.35) | 0.90 (0.26) | 0.1666 | 1.06 (0.42) | 0.97 (0.34) | 0.1604 |
| Urea [mg/dL]* | 33.5 (21.17) | 28.8 (12.15) | 0.0735 | 33.5 (13.64) | 33.0 (14.97) | 0.8335 |
| Urea/Creatinine* | 34.8 (12.84) | 32.8 (12.19) | 0.2769 | 32.8 (10.51) | 34.6 (12.17) | 0.3348 |
| CRP determined | 24 (30.8%) | 48 (29.3%) | 0.8113 | 54 (96.4%) | 91 (88.3%) | 0.1406~ |
| PCT determined | 21 (26.9%) | 15 (9.15%) | 0.0003 | 12 (21.4%) | 7 (6.80%) | 0.0066 |
| GCS | 12 (4–15) | 13 (8–15) | 0.0109# | 11 (3–14) | 14 (7–15) | 0.0127# |
| EK | 43 (55.1%) | 52 (31.7%) | 0.0005 | 22 (39.3%) | 22 (21.4%) | 0.0158 |
| SOFA Total* | 9 (7–11) | 6 (4–9) | <.0001# | 9 (6–10) | 6 (4–9) | 0.0002# |
| SOFA Respiratory* | | | 0.0054~ | | | 0.7455~ |
| 0 | 0 (0%) | 6 (3.66%) | | 1 (1.79%) | 3 (2.91%) | |
| 1 | 9 (11.5%) | 24 (14.6%) | | 6 (10.7%) | 8 (7.77%) | |
| 2 | 14 (17.9%) | 49 (29.9%) | | 18 (32.1%) | 43 (41.7%) | |
| 3 | 33 (42.3%) | 65 (39.6%) | | 27 (48.2%) | 41 (39.8%) | |
| 4 | 22 (28.2%) | 18 (11.0%) | | 4 (7.14%) | 8 (7.77%) | |
| SOFA Cardiovascular | | | 0.0004 | | | 0.0026 |
| 0 | 14 (17.9%) | 56 (34.1%) | | 7 (12.5%) | 21 (20.4%) | |
| 1 | 30 (38.5%) | 75 (45.7%) | | 8 (14.3%) | 38 (36.9%) | |
| 3 | 8 (10.3%) | 13 (7.93%) | | 13 (23.2%) | 16 (15.5%) | |
| 4 | 26 (33.3%) | 20 (12.2%) | | 28 (50.0%) | 28 (27.2%) | |
| SOFA Coagulation* | | | <.0001~ | | | 0.3708~ |
| 0 | 13 (16.7%) | 72 (43.9%) | | 28 (50.0%) | 60 (58.3%) | |
| 1 | 23 (29.5%) | 49 (29.9%) | | 18 (32.1%) | 28 (27.2%) | |
| 2 | 38 (48.7%) | 37 (22.6%) | | 7 (12.5%) | 12 (11.7%) | |
| 3 | 4 (5.13%) | 5 (3.05%) | | 3 (5.36%) | 1 (0.97%) | |
| SOFA Renal* | | | 0.5654~ | | | 0.5502~ |
| 0 | 63 (80.8%) | 141 (86.0%) | | 42 (75.0%) | 79 (76.7%) | |
| 1 | 14 (17.9%) | 21 (12.8%) | | 12 (21.4%) | 21 (20.4%) | |
| 2 | 1 (1.28%) | 1 (0.61%) | | 2 (3.57%) | 1 (0.97%) | |
| SOFA Hepatic | | | 0.0107~ | | | 0.5226~ |
| 0 | 56 (71.8%) | 139 (84.8%) | | 47 (83.9%) | 93 (90.3%) | |

| | Cohort 1 (N=242) | | | Cohort 2 (N=159) | | |
|---|---|---|---|---|---|---|
| | Sepsis (N=78) | No sepsis (N=164) | | Sepsis (N=56) | No sepsis (N=103) | |
| | Mean (SD) Median (IQR) n (%) | Mean (SD) Median (IQR) n (%) | p-value | Mean (SD) Median (IQR) n (%) | Mean (SD) Median (IQR) n (%) | p-value |
| 1 | 13 (16.7%) | 21 (12.8%) | | 7 (12.5%) | 8 (7.77%) | |
| 2 | 7 (8.97%) | 4 (2.44%) | | 2 (3.57%) | 2 (1.94%) | |
| 3 | 2 (2.56%) | 0 (0%) | | 0 (0%) | 0 (0%) | |
| SOFA Neuro | | | 0.0720 | | | 0.0895 |
| 0 | 24 (30.8%) | 73 (44.5%) | | 12 (21.4%) | 43 (41.7%) | |
| 1 | 12 (15.4%) | 20 (12.2%) | | 10 (17.9%) | 17 (16.5%) | |
| 2 | 6 (7.69%) | 22 (13.4%) | | 8 (14.3%) | 9 (8.74%) | |
| 3 | 13 (16.7%) | 20 (12.2%) | | 7 (12.5%) | 13 (12.6%) | |
| 4 | 23 (29.5%) | 29 (17.7%) | | 19 (33.9%) | 21 (20.4%) | |
| SIRS average λ | 1.70 (0.67) | 1.28 (0.73) | <.0001 | 1.74 (0.65) | 1.36 (0.71) | 0.0012 |
| SIRS count C | 14.5 (10.44) | 17.7 (11.08) | 0.0303 | 13.9 (6.57) | 15.3 (9.20) | 0.2763 |
| SIRS Δ | 0.5 (1.36) | 0.2 (1.36) | 0.1074 | 0.9 (1.16) | 0.8 (1.21) | 0.5596 |

## The statistics of outcomes measured at onset time point

| | Cohort 1 (N=242) | | | Cohort 2 (N=159) | | |
|---|---|---|---|---|---|---|
| | Sepsis (N=78) | No sepsis (N=164) | | Sepsis (N=56) | No sepsis (N=103) | |
| | Mean (SD) | Mean (SD) | p-value | Mean (SD) | Mean (SD) | p-value |
| Creatinine [mg/dL]* | 1.04 (0.49) | 0.81 (0.37) | 0.00027 | 1.07 (0.53) | 0.92 (0.49) | 0.088 |
| Urea [mg/dL]* | 59.9 (35.5) | 33 (17.6) | <.0001 | 57.4 (35.3) | 41.2 (25) | 0.0032 |
| Lactate [mmol/L]* | 1.2 (0.76) | 0.83 (0.33) | <.0001 | 1.1 (0.56) | 0.88 (0.38) | 0.0069 |

**Table S3:** Statistical Summary of Features and Outcomes in Sepsis Analysis. This table presents a statistical summary of the features and outcomes used in the analysis of sepsis. The p-values referred to the significance of differences between the sepsis and non-sepsis groups. Features that contain missing data are marked with an asterisk (*). The method of p-value calculation varies depending on the data type: a t-test is used for continuous features, while a Chi-squared ($\chi^2$) test is used for categorical features. P-values calculated using the Mann-Whitney-Wilcoxon test are denoted by a hash (#), and those calculated using Fisher's exact test are indicated by a tilde (~).

| Pathway ID | Pathway name | FDR value |
|---|---|---|
| GO:0005244 | voltage-gated monoatomic ion channel activity | 0.003115691 |
| GO:0022832 | voltage-gated channel activity | 0.003115691 |
| GO:0022832 | gated channel activity | 0.007911141 |
| GO:0022839 | monoatomic ion gated channel activity | 0.007911141 |
| GO:0007268 | chemical synaptic transmission | 0.025878397 |
| GO:0098916 | anterograde trans-synaptic signaling | 0.025878397 |
| GO:0099537 | trans-synaptic signaling | 0.025878397 |
| GO:0009887 | animal organ morphogenesis | 0.025878397 |
| GO:0030001 | metal ion transport | 0.025878397 |
| GO:0006469 | negative regulation of protein kinase activity | 0.025878397 |
| GO:0007423 | sensory organ development | 0.032789873 |
| GO:0099536 | synaptic signaling | 0.033531818 |
| GO:0033673 | negative regulation of kinase activity | 0.046850028 |

**Table S4:** The enriched pathways using top selected genes of MTLComb model from Schizophrenia analysis.